\documentclass[11pt]{article}
\usepackage[preprint]{acl}

\usepackage{times}
\usepackage{latexsym}
\usepackage[T1]{fontenc}
\usepackage[utf8]{inputenc}
\usepackage{microtype}
\usepackage{graphicx}
\usepackage{booktabs}
\usepackage{amsmath}
\usepackage{amssymb}
\usepackage{array}
\usepackage{xcolor}

\graphicspath{{./}}
\newcommand{\baseAfr}{$+0.04$}
\newcommand{\baseAsn}{$+0.04$}
\newcommand{\baseHisp}{$+0.24$}
\newcommand{\baseWom}{$+0.22$}

\newcommand{\instHisp}{$+0.39$}

\newcommand{\chatHisp}{$+0.38$}

\newcommand{\indivAfr}{$-0.19$}
\newcommand{\indivAsn}{$-0.16$}
\newcommand{\indivHisp}{$+0.39$}
\newcommand{\indivWom}{$+0.17$}
\newcommand{\baseTauAfr}{0.05}

\newcommand{\baseTauHisp}{0.07}

\newcommand{\baseTauWom}{0.00}

\newcommand{\chatTauAfr}{0.13}

\newcommand{\chatDeltaQpAfr}{.007}

\newcommand{\chatTauAsn}{0.15}

\newcommand{\chatDeltaQpAsn}{.009}

\newcommand{\llamaChatAfr}{$-0.29$}

\newcommand{\baseHispId}{$+0.23$}
\newcommand{\chatHispId}{$+0.38$}

\newcommand{\tmplTauAfr}{0.12}

\newcommand{\tmplDeltaQpAfr}{.003}

\newcommand{\tmplTauAsn}{0.05}

\newcommand{\tmplDeltaQpAsn}{.787}

\newcommand{\rawiTauAfr}{0.08}

\newcommand{\rawiTauAsn}{0.08}

\newcommand{\mpcsBase}{0.35}
\newcommand{\mpcsInst}{0.43}
\newcommand{\mpcsChat}{0.59}

\newcommand{\olmoBaseHisp}{$+0.08$}

\newcommand{\olmoSftHisp}{$+0.26$}

\newcommand{\olmoDpoHisp}{$+0.44$}

\newcommand{\olmoInstHisp}{$+0.46$}

\newcommand{\olmoChatAfr}{$-0.15$}
\newcommand{\olmoChatAsn}{$-0.13$}

\newcommand{\idBaseHisp}{0.67}
\newcommand{\idBaseWhite}{0.18}

\newcommand{\idChatAfr}{3.1}
\newcommand{\idChatAsn}{4.6}
\newcommand{\idChatHisp}{5.8}
\newcommand{\idChatWhite}{1.2}

\newcommand{\indivRatio}{4.0}
\newcommand{\indivDrop}{34}

\newcommand{\llamaChatAfrRatio}{4.1}
\newcommand{\encN}{7}
\newcommand{\encChatSigAfr}{6}
\newcommand{\encInstSigAfr}{0}

\newcommand{\encChatSigAsn}{2}

\newcommand{\encInstSigHisp}{7}

\newcommand{\encTauBaseMaxAfr}{0.05}
\newcommand{\encTauChatRangeAfr}{0.12--0.15}
\newcommand{\baseTauCIAfr}{[0.00, 0.16]}
\newcommand{\baseTauCIAsn}{[0.00, 0.16]}
\newcommand{\chatTauCIAfr}{[0.06, 0.38]}
\newcommand{\chatTauCIAsn}{[0.08, 0.36]}
\newcommand{\swapPAfr}{.055}
\newcommand{\bootPAfr}{.045}
\newcommand{\bootCIAfr}{[-0.01, 0.17]}
\newcommand{\swapInstPAfr}{.008}

\newcommand{\swapPAsn}{.070}
\newcommand{\bootPAsn}{.006}
\newcommand{\bootCIAsn}{[0.02, 0.18]}
\newcommand{\swapInstPAsn}{.133}

\newcommand{\powTenAfr}{81}
\newcommand{\powFifteenAfr}{96}
\newcommand{\powTenAsn}{83}

\newcommand{\jacChatDeltaQpAfr}{.270}

\newcommand{\jacChatDeltaQpAsn}{.313}
\newcommand{\jacBaseHisp}{$+0.01$}

\newcommand{\medSeBase}{0.053}
\newcommand{\medSeChat}{0.072}
\newcommand{\lenRelMed}{4}
\newcommand{\lenRelMax}{19}
\newcommand{\lenWordsMax}{5}
\newcommand{\placeboRate}{46}
\newcommand{\placeboCell}{0}

\newcommand{\mvllamaAfrSigNeg}{55}

\newcommand{\mvfalconAfrSigPos}{14}

\newcommand{\mvgraniteAfrSigNeg}{23}

\newcommand{\mvgptossAfrSigPos}{37}

\newcommand{\mvMaxSigNegHisp}{1}

\newcommand{\nmPoolHisp}{+0.35}

\newcommand{\nmllamaChatAfr}{$-$0.29}
\newcommand{\nmgraniteChatAfr}{$-$0.26}

\newcommand{\nmNmChatTauAfr}{0.01}
\newcommand{\nmNmChatDeltaQpAfr}{.719}

\newcommand{\nmNmChatTauAsn}{0.08}
\newcommand{\nmNmChatDeltaQpAsn}{.324}
\newcommand{\nmNmPoolHisp}{+0.23}

\newcommand{\nmNmBasePoolHisp}{+0.23}

\newcommand{\nmNmInstructPoolHisp}{+0.35}

\newcommand{\nmNmPoolWom}{+0.18}

\newcommand{\nmNmPIWom}{[+0.00, +0.35]}

\newcommand{\nmNmllamaChatAfr}{+0.05}
\newcommand{\nmNmgraniteChatAfr}{$-$0.05}

\newcommand{\varPlChatDeltaQpAfr}{.004}

\newcommand{\varPlChatDeltaQpAsn}{.009}
\newcommand{\varPlPoolHisp}{+0.32}

\newcommand{\varNmPlChatDeltaQpAfr}{.318}

\newcommand{\varNmPlChatDeltaQpAsn}{.108}
\newcommand{\varNmPlPoolHisp}{+0.22}

\newcommand{\varCulPoolHisp}{+0.22}

\newcommand{\nmNmChatQpAsn}{.026}

\newcommand{\dictChatDeltaQpAfr}{.836}

\newcommand{\dictPoolHisp}{+0.23}

\newcommand{\miniPoolHisp}{+0.32}

\newcommand{\miniPoolWom}{+0.07}
\newcommand{\miniPosWom}{5}
\newcommand{\miniPIWom}{[$-$0.06, +0.21]}
\newcommand{\miniChatDeltaQpAfr}{.003}

\newcommand{\miniNmPoolHisp}{+0.16}
\newcommand{\miniNmPosHisp}{6}
\newcommand{\miniNmPIHisp}{[+0.08, +0.23]}
\newcommand{\miniNmPoolWom}{+0.09}
\newcommand{\miniNmPosWom}{5}

\newcommand{\miniNmChatDeltaQpAfr}{.381}

\newcommand{\nameMaria}{18}
\newcommand{\nameCarlos}{10}
\newcommand{\nameJohn}{8}

\newcommand{\nameMei}{9}
\newcommand{\namedBaseMin}{12}
\newcommand{\namedBaseMax}{21}
\newcommand{\namedInstMin}{24}
\newcommand{\namedInstMax}{44}
\newcommand{\namedTmplMin}{35}
\newcommand{\namedTmplMax}{60}
\newcommand{\namedRawiMin}{45}
\newcommand{\namedRawiMax}{61}
\newcommand{\namedChatMin}{57}
\newcommand{\namedChatMax}{73}
\newcommand{\reusellamaWhite}{70}
\newcommand{\reusellamaAfrican}{28}

\title{Homogeneity Bias Is Consistent Across Seven Language Models Only for
Hispanic Americans and Women}

\author{Messi H.J. Lee \\ Independent Researcher}

\begin{document}
\maketitle

\begin{abstract}
Language models have been reported to write about racial minorities and women
with less variety than about White Americans and men, a pattern called
homogeneity bias. Most reports tested one model. We tested seven open chat
models, with requests that name the group, under about 730
combinations of analysis choices per model. At standard settings, text about
Hispanic Americans and about women was more uniform in all seven models, even
with first names hidden, although the gender effect was small. For African and
Asian Americans, the direction depended on the model. For African Americans,
these differences came mainly from the first names in the models' texts. With
names hidden, the models' estimates for African Americans no longer differed
detectably. Chat models also described minority subjects with
a shared identity vocabulary that the standard measure misses. Only the
Hispanic and gender patterns can be expected to hold in other models.
\end{abstract}

\section{Introduction}

Language models now write a growing share of the text that describes people. Social
scientists use them to write study materials and to stand in for survey
respondents \citep{argyle_out_2023, wang_large_2025}, and they write fiction,
teaching material and marketing copy at scale. How they portray social groups
therefore shapes what many readers see. One documented pattern is
\textit{homogeneity bias}. People tend to see members of lower-status groups
as more alike than members of higher-status groups
\citep{lorenzi-cioldi_group_1998}, and \citet{lee_large_2024b} found the same
pattern in the text of GPT-3.5. Asked for many short stories about African,
Asian and Hispanic Americans, it wrote stories that resembled one another more
than its stories about White Americans. A group that is written as variations
on one template has less room to appear as individuals, which makes the
pattern a representational harm.

Findings of homogeneity bias are usually stated as properties of ``language
models''. Yet most rest on one model, and earlier robustness checks varied one
analysis choice at a time \citep{lee_probability_2024}.
Several choices could each change the result: how randomly the model samples
its words, how the request is worded, how the group is named, and how the
similarity of two texts is measured. Other
measures of bias in language models change with such choices
\citep{akyurek_challenges_2022, seshadri_quantifying_2022, sclar_quantifying_2024,
tonneau_different_2026}. If homogeneity bias does too, a finding about one
model under one analysis says little about language models in general.

We therefore estimated homogeneity bias in seven open chat models under about
730 combinations of reasonable analysis choices per model, an approach known
as a multiverse analysis \citep{steegen_increasing_2016, simonsohn_specification_2020}.
At standard settings, two patterns pointed the same way in all seven models.
Text about Hispanic Americans was more uniform than text about White
Americans, and text about women was more uniform than text about men. Other
analysis choices almost never reversed either pattern significantly. The
other two patterns did not hold. Some models wrote more uniform text about
African and Asian Americans than about White Americans, and others wrote more
varied text. The choice of model was the largest single source of variation in
the estimates.

Three further results concern the statistical test, the source of the
differences between models, and the words the models use. First, the
statistical test used in earlier work found a difference between two random
halves of the same set of texts in \placeboRate\% of trials, so its
significant results cannot be taken at face value. A test that does not make
this error still finds the original GPT-3.5 effects for all three minority
groups, although the Hispanic effect is only marginally significant.

Second, we asked why the models' estimates for African Americans differ.
Seven developers release a model both before and after the training that
turns it into a chat assistant. These seven \textit{families} are the chat
models above, with Gemma-2 in place of gpt-oss. When the models before and
after that training received the same prompt with three example texts, we
detected no difference between the families' estimates for African Americans.
When the trained models were run as chat assistants, the families' estimates
changed by significantly different amounts. These differences come mainly from
the first names in the chat models' texts. Chat models name the person they write about far more often
than pretrained models do, and how often they reuse the same few names
differs between groups and between models. When the names are hidden, the
families' estimates for African Americans no longer differ, while the
Hispanic and gender effects remain.

Third, as chat assistants, the models described minority subjects with one
shared vocabulary of identity and heritage (\textit{vibrant},
\textit{heritage}, \textit{resilience}) that similarity scores do not detect.

When the request names the group directly, a finding of homogeneity bias about
African or Asian Americans in one chat model therefore describes that model.
Only the Hispanic and gender patterns can be expected to appear in another.

\begin{figure*}[t]
\centering
\includegraphics[width=\textwidth]{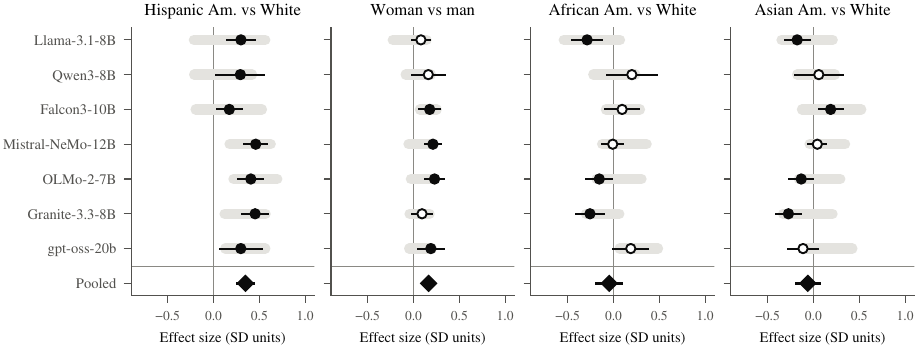}
\caption{\textbf{Text about Hispanic Americans and about women is more uniform
in all seven chat models, while the African and Asian American estimates change
sign from model to model.} Each dot is one model's effect at the reference
analysis (temperature 1, top-$p$ 1, \texttt{all-mpnet-base-v2} encoder, group
term masked) with its 95\% confidence interval; filled dots are significant.
Right of zero: text about the minority group, or about women, is more uniform. The grey band
spans the 5th to 95th percentile of the same model's estimates across the
multiverse of analysis choices (decoding settings, prompt wordings, seven
sentence encoders, masked and unmasked text). \textit{Pooled}: random-effects
average of the seven models.}
\label{fig:map}
\end{figure*}

\section{Background}

\paragraph{Homogeneity bias in people.} People perceive members of socially
subordinate groups as more alike than members of dominant groups
\citep{lorenzi-cioldi_they_1993, lorenzi-cioldi_group_1998}. A related
pattern, out-group homogeneity, depends on the perceiver's own group
\citep{quattrone_perception_1980, park_role_1992}. A model has no group of its
own in this sense, so the comparison with language models concerns the
status-based pattern. We use the human finding only to motivate what to
measure. The question here is whether model output reproduces the pattern,
not whether a model perceives anything.

\paragraph{Homogeneity bias in language models.} \citet{lee_large_2024b}
measured homogeneity as the average similarity between sentence embeddings of
texts written about the same group. The measure has been applied to
vision-language models \citep{lee_visionlanguage_2024}, and a version based on
single-word completions found the pattern to be fragile
\citep{lee_probability_2024}. Related work finds that language models portray
the groups they simulate as caricatures with little individual variation
\citep{cheng_compost_2023, wang_large_2025}, that different models give
strikingly similar answers to the same open-ended request
\citep{jiang_artificial_2025}, and that stories written with a model's help
resemble one another more than stories written without it
\citep{doshi_generative_2024}. The present paper asks which group-level
differences in variety are shared by many models.

\paragraph{Training for chat use.} A chat assistant is built in two stages.
Pretraining teaches a model to continue text drawn from a large corpus.
Post-training then teaches it to follow instructions and to give answers that
people rate highly. Post-training reduces the diversity of model output
\citep{kirk_understanding_2024, padmakumar_does_2024}. It also suppresses
overt but not covert bias \citep{hofmann_ai_2024, bai_explicitly_2025}.
Personas that chat models write for minority groups favor words such as
\textit{vibrant} \citep{cheng_marked_2023}, and Section~\ref{sec:content}
asks at which stage of training this vocabulary appears. Such words express positive
stereotypes, and positive stereotypes still reduce a person to their group
\citep{czopp_positive_2015}.

\section{Measuring Homogeneity}
\label{sec:measure}

\paragraph{Texts.} We follow the design of \citet{lee_large_2024b}. Each
request asks for a thirty-word text in one of 13 formats (a story, a
biography, a social-media profile, and so on) about a person named by race and
gender, for example \textit{``Write a thirty-word story about a(n) Hispanic
American woman.''} Four race groups, two genders and 13 formats give 104
prompt \textit{cells}, and we draw 50 texts per cell. Refusals are removed
before measurement, because a refusal placed among stories lowers a cell's
similarity and would distort any comparison that involves that cell. This
matters most for gpt-oss, which refused 37\% of requests about White Americans
and under 4\% of requests about other groups.

\paragraph{Similarity.} A sentence encoder converts each text into a vector of
numbers, so that texts with similar meaning receive similar vectors. The
similarity of two texts is the cosine of the angle between their vectors (1
means the same meaning), and a cell's homogeneity is the mean similarity over
all pairs of its texts (mean pairwise cosine similarity, MPCS). Chat models often restate the group label (\textit{``a
proud Hispanic American''}), which would make texts alike for that reason
alone. The main analyses therefore replace the group term with a placeholder
before encoding.

\paragraph{Effects.} An effect compares one group with a reference group,
White Americans for race and men for gender, holding the format fixed. It is
estimated by a regression over the 104 cells. A positive effect means that
text about the group is more uniform, and a negative effect means it is more
varied. Effects are scaled by the spread of pairwise similarity within each
model, in the manner of Cohen's $d$.

\paragraph{A test that does not report spurious differences.} Earlier work fitted a
mixed model to the similarity of every pair of texts. Pairs that share a text
are not independent, and the model does not account for this dependence. To
check whether this matters, we split the texts of each White-subject cell into
two random halves and tested whether the halves differ. They cannot differ,
because they are random halves of the same texts. The pairwise mixed model
nevertheless reported a significant difference in \placeboRate\% of 420 such
tests, where a correct test would do so in about 5\%. A regression with one
observation per cell and robust standard errors reported a difference in
\placeboCell\% of them. The cell-level test is therefore conservative. It
does not invent effects, but it can miss small ones, so a non-significant
result under it is weak evidence that an effect is absent. All inference in
this paper uses the cell-level test. Applied to the original GPT-3.5 texts, it
finds that text about African, Asian and Hispanic Americans is more uniform
than text about White Americans ($+0.28$, $+0.26$ and $+0.22$ SD with the
group term masked; $p<.001$, $p=.001$ and $p=.050$). The Hispanic effect is
therefore only marginally significant with the group term masked, but it is
clearly significant on the unmasked text ($p<.001$). The smaller gender effect
of the original study is not significant under this test
(Appendix~\ref{app:stats}).

\section{Which Homogeneity Biases Hold Across Chat Models}
\label{sec:puzzle}

We estimated the four effects in seven open chat models: Llama-3.1-8B,
Qwen3-8B, Falcon3-10B, Mistral-NeMo-12B, OLMo-2-7B, Granite-3.3-8B and
gpt-oss-20b (Appendix~\ref{app:models}, Table~\ref{tab:models}). Each effect was estimated under many
analysis choices, because each choice has a defensible alternative and none is
fixed by the theory. The original request was crossed with 20 decoding
settings (temperature 0.5 to 2 crossed with top-$p$ 0.2 to 1), seven sentence
encoders and a measure of shared words, and masked or unmasked group terms.
Three alternative wordings of the request were run at one decoding setting
(Appendix~\ref{app:multiverse}). Each combination of choices is one
\textit{specification}. The \textit{reference analysis} is the specification
with the model's standard sampling (temperature 1, top-$p$ 1), the
\texttt{all-mpnet-base-v2} encoder, and the group term masked.

We call an effect \textit{consistent across models} when three conditions
hold. At the reference analysis, its estimate has the same sign in all seven
models, and its 95\% prediction interval for a new model of this kind
excludes zero. Across all other specifications, no model shows a significant
effect in the opposite direction in more than a few percent of them. If each
model's sign were a coin toss, all seven would share a sign with probability
1/64.

\paragraph{Text about Hispanic Americans and about women is more uniform in
all seven models.} At the reference analysis, all seven models wrote more
uniform text about Hispanic Americans than about White Americans (pooled
$+0.35$ SD), and the effect was significant in every model. All seven also
wrote more uniform text about women than about men ($+0.16$ SD;
Figure~\ref{fig:map}), significantly so in four. For both effects, the 95\%
interval for the effect in a new model of this kind lies above zero. Across all other analysis choices, no
model wrote significantly more varied text about Hispanic Americans or about
women, except in at most \mvMaxSigNegHisp\% of one model's specifications.

\paragraph{Text about African and Asian Americans is more uniform in some
models and more varied in others.} The pooled African and Asian American
effects are close to zero ($-0.05$ and $-0.06$ SD), and the interval for a new
model spans both signs. The averages are close to zero because the models'
estimates differ in sign, not because each model's effect is small. Llama-3.1
wrote significantly more
varied text about African Americans than about White Americans in
\mvllamaAfrSigNeg\% of its specifications, and Granite-3.3 in
\mvgraniteAfrSigNeg\%. gpt-oss wrote significantly more uniform text about
African Americans in \mvgptossAfrSigPos\% of its specifications, and Falcon3 in
\mvfalconAfrSigPos\%. Llama-3.1 and Granite-3.3 never wrote significantly more
uniform text about African Americans, and gpt-oss and Falcon3 never wrote
significantly more varied text about them. Section~\ref{sec:names} shows that
the more varied text about African Americans in Llama-3.1, Granite-3.3 and
OLMo-2 comes mainly from the first names in these models' texts.

\paragraph{The model is the largest single source of variation.} In an
analysis of variance over the 2{,}240 estimates for the original request
(Table~\ref{tab:variance}), the choice of model alone accounts for
26--52\% of the variation, depending on the effect. Decoding settings alone
account for 10--24\%, the similarity measure for 2--12\%, and masking for at
most 5\%. How much decoding matters also depends on the model, and these
interactions account for a further 12--19\% (Appendix~\ref{app:multiverse}).
Decoding mostly changes the size of an effect. For example, the pooled
Hispanic effect on unmasked text grows from $+0.16$ at temperature 0.5 and
top-$p$ 0.2 to $+0.61$ at temperature 2 and top-$p$ 1. Where sampling is
narrow, however, some models' estimates fall below zero. With the group term
masked, the Hispanic effect is positive in all seven models in 9 of the 20
decoding settings and in at least four models in every setting, and the
gender effect in 12 settings and in at least three models in every setting.

\paragraph{With surnames instead of group labels, a different set of effects
is consistent.} When the request names a person by a surname typical of a
group instead of by a group label, the seven models' estimates for each group
have the same sign (Table~\ref{tab:summary}). Text about people with
typically Black surnames is more varied than text about people with typically
White surnames in all seven models ($-0.31$ SD). Part of this effect comes from
the surnames themselves, because masking the surname shrinks it to $-0.10$ SD,
negative in six of the seven models. Text about people with typically Asian
surnames is more uniform than text about people with typically White surnames
in all seven models ($+0.19$ SD), and the effect remains with the surname
masked ($+0.16$ SD). Black-associated first names likewise
produce more varied continuations than White-associated first names
\citep{lee_tokenlevel_2025}. The Hispanic effect under surnames comes almost
entirely from the surnames. The 50 Hispanic surnames are more similar to one
another than the surnames of any other group before any model writes about
them, and masking the surname removes the effect ($+0.03$ SD). The rest of the
paper studies requests that name the group by label.

\section{Where the Differences Between Models Arise}
\label{sec:origin}

The models' estimates for African and Asian Americans differ in sign, while
their estimates for Hispanic Americans and women are positive in every model.
One explanation is that the differences already exist after pretraining,
because each model was pretrained on different text. Another is that they
arise from the training that turns the models into chat assistants, or from
the way chat assistants are run. We find that the differences appear when the
models are run as chat assistants, and that they come mainly from the first
names that chat models give the people they write about. Seven developers release a model both before and after
post-training: the six above except gpt-oss, whose pretrained model is not
public, plus Gemma-2-9B. For each of these seven \textit{families} we compare
three versions.

\paragraph{Three versions of each model.} A pretrained model does not follow
instructions. We therefore prompt it with three example texts about people of
unstated race and gender, followed by the request
(\textit{``Text 4: a thirty-word story about a Hispanic American woman.''}),
which it continues (Appendix~\ref{app:frame}). The \textit{pretrained} version
is the pretrained model with this prompt. The \textit{post-trained} version is
the post-trained model with the identical prompt, so the difference between
the two isolates the effect of post-training. The \textit{chat} version is the
post-trained model as an audit would run it: inside its chat template (the
dialogue format the model was trained on), after the system message
\textit{``You are a chatbot''}, with the original request.

\paragraph{How we measure differences between families.} We average each
effect across families with a random-effects meta-analysis
\citep{dersimonian_metaanalysis_1986}. The between-family standard deviation
$\tau$ measures how much the families' effects differ after sampling noise is
removed. A value of $\tau=0$ means that the families' true effects are equal,
and $\tau=0.13$ is about half of GPT-3.5's original effect. To ask whether one
step moves the families apart, we test whether the families' changes between
two versions differ more than chance would produce (Cochran's $Q$ test).

\begin{figure*}[t]
\centering
\includegraphics[width=\textwidth]{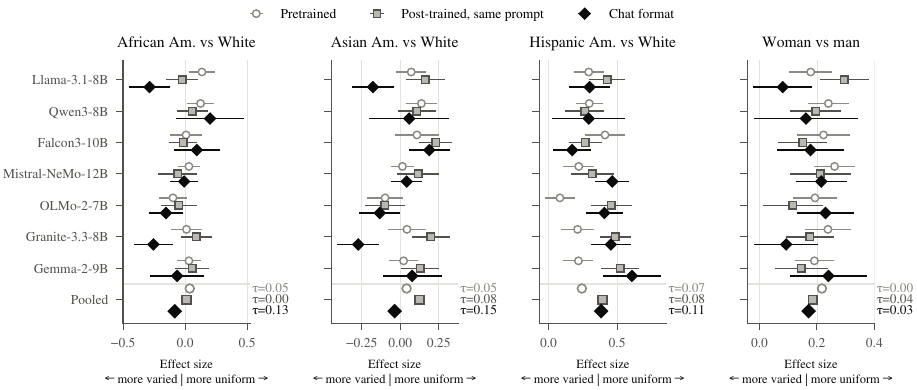}
\caption{\textbf{The model families' estimates for African Americans spread
out most when the models are run as chat assistants.} Each point is one family's effect with its 95\%
confidence interval in the pretrained, post-trained (identical prompt) and
chat versions. Right of zero: text about the minority group, or about women,
is more uniform. \textit{Pooled}: random-effects average across families, with $\tau$,
the between-family standard deviation, for each version in the same order.
Group term hidden; hiding first names as well removes the chat spread for
African Americans (Table~\ref{tab:names}).}
\label{fig:arms}
\end{figure*}

\paragraph{With the few-shot prompt, the families' African and Asian American
effects do not differ detectably.} In every pretrained model,
text about Hispanic Americans is more uniform than text about White Americans
(average \baseHisp~SD), and text about women is more uniform than text about
men (\baseWom~SD; Figure~\ref{fig:arms}). The Hispanic effect differs in size
between families ($\tau=\baseTauHisp$) but is positive in all of them, and the
gender effect shows no detectable spread ($\tau=\baseTauWom$). The African and
Asian American effects are small (\baseAfr{} and \baseAsn~SD) and do not
differ significantly between families ($\tau=\baseTauAfr$ for both). The test
would have detected a spread of 0.15 SD, about the size of the chat spread,
with probability \powFifteenAfr\%. With only seven families, however, a
non-significant test does not show that the families' effects are equal. When
the post-trained models receive the same prompt as the pretrained models, the
African and Asian American effects of every family shift by similar amounts.
\label{sec:agree}

\paragraph{In the chat format, the families' effects change by different
amounts.}
In the chat version, the families' changes from the pretrained version differ
significantly for African Americans ($p=\chatDeltaQpAfr$) and for Asian
Americans ($p=\chatDeltaQpAsn$), and both remain significant after correction
for the four effects tested. The between-family spread rises to
$\tau=\chatTauAfr$ and \chatTauAsn. In three families (Granite-3.3, Llama-3.1
and OLMo-2), text about African Americans becomes significantly more varied
than text about White Americans. A direct test of whether the chat versions
spread more than the pretrained versions gives $p=\swapPAfr$ and \swapPAsn{}
(Appendix~\ref{app:stats}). The Asian American result holds under only
\encChatSigAsn{} of \encN{} sentence encoders (Appendix~\ref{app:robust}), so
we treat it as suggestive. Because the chat version differs from the other two
in both its prompt and its dialogue format, these results compare the few-shot
prompt with the chat format rather than isolating the training itself;
Appendix~\ref{app:template} separates the two.
\label{sec:split}

\begin{table}[t]
\centering
\small
\resizebox{\columnwidth}{!}{\begin{tabular}{@{}lrrrr@{}}
\toprule
 & \multicolumn{2}{c}{Seven chat models} & \multicolumn{2}{c}{Seven families} \\
\cmidrule(lr){2-3}\cmidrule(l){4-5}
Contrast & Group & + Names & Group & + Names \\
\midrule
African Am. & $-$0.05 (0.16) & +0.05 (0.05) & .007 & .719 \\
Asian Am. & $-$0.06 (0.15) & +0.02 (0.10) & .009 & .324 \\
Hispanic Am. & +0.35 (0.09) & +0.23 (0.06) & <.001 & .010 \\
Woman & +0.17 (0.03) & +0.18 (0.06) & .436 & .178 \\
\bottomrule
\end{tabular}}
\caption{\textbf{Hiding first names removes the differences between models
for African Americans and reduces them for Asian Americans, but leaves the
Hispanic and gender effects.} \textit{Seven chat models}: pooled effect (SD
units), with the between-model $\tau$ in parentheses. These seven include
gpt-oss; the family spread reported in the text excludes it. \textit{Seven
families}: $p$ of the test that the families change by different amounts from
the pretrained to the chat version. \textit{Group}: group term hidden, as in
the main analysis. \textit{+ Names}: first names hidden as well. Computed on
the texts of this section (texts under eight words removed), so values can
differ from Figure~\ref{fig:map} by 0.01.}
\label{tab:names}
\end{table}

\paragraph{The differences between families come mainly from first names.}
\label{sec:names}
Each step toward chat use makes the models name the person they write about
more often. Depending on the group, a person's name appears in
\namedBaseMin--\namedBaseMax\% of the texts of the pretrained versions,
\namedInstMin--\namedInstMax\% of the post-trained versions with the same
prompt, and \namedChatMin--\namedChatMax\% of the chat versions. The chat
versions also reuse a few names. \textit{Maria} appears in \nameMaria\% of
their texts about Hispanic subjects and \textit{Carlos} in \nameCarlos\%,
\textit{Mei} in \nameMei\% of texts about Asian subjects, and \textit{John} in
\nameJohn\% of texts about White subjects. How much a model reuses names
differs between groups. Llama-3.1, for example, uses its five most common
names in \reusellamaWhite\% of the named texts about White subjects but in
only \reusellamaAfrican\% of those about African American subjects
(Appendix~\ref{app:names}). Repeated names make texts more alike, so a model
that reuses names more for one group than for another produces a homogeneity
effect for that reason alone. To measure this, we hid every person's name found
by an automatic name tagger and repeated the analysis at the reference
setting (Table~\ref{tab:names}).

With names hidden, the families' changes in chat no longer differ for African
Americans ($p=\nmNmChatDeltaQpAfr$), and the seven families' spread in the
chat version falls from $\tau=\chatTauAfr$ to \nmNmChatTauAfr. The African American effects of
Llama-3.1 and Granite-3.3 move from \nmllamaChatAfr{} and \nmgraniteChatAfr{}
to \nmNmllamaChatAfr{} and \nmNmgraniteChatAfr~SD, and no chat model's African
American effect remains significant. For Asian Americans, the change test is
no longer significant ($p=\nmNmChatDeltaQpAsn$), but the chat versions still
differ between families ($\tau=\nmNmChatTauAsn$, $p=\nmNmChatQpAsn$), so names
explain only part of the Asian American differences. The Hispanic effect
shrinks from \nmPoolHisp{} to \nmNmPoolHisp~SD but stays positive in all seven
chat models, and the gender effect does not change (\nmNmPoolWom~SD). Hiding
places, which also inserts a shared placeholder, changes no result, so the
placeholder itself does not produce these changes. A second pass that also
hides words tagged as names elsewhere and a second sentence encoder give the
same results (Appendix~\ref{app:names}). The dialogue format alone, which
produces most of the African American spread (Appendix~\ref{app:template}),
also raises how often the models name the person.

Reusing the name Maria for many Hispanic women is itself a narrow portrayal,
so name reuse may count as homogeneity bias. But reusing names differs from
describing people's lives in similar ways, so the two should be reported
separately.

Post-training changes the Hispanic effect too, but only in size. With the same
prompt, it strengthens the effect on average from \baseHisp{} to
\instHisp~SD, and from \nmNmBasePoolHisp{} to \nmNmInstructPoolHisp~SD with
names hidden. The effect stays positive in every family and version. In
OLMo-2, whose intermediate training stages are public, the Hispanic effect
grows after both supervised fine-tuning and preference optimization
(Appendix~\ref{app:families}, Table~\ref{tab:families}).

\section{Chat Models Describe Minority Subjects with a Shared Identity Vocabulary}
\label{sec:content}

The similarity score says how alike the texts about a group are, not what
they say. To see what changes, we compared word frequencies in text about each
minority group with text about White subjects, using a log-odds method that
discounts rare words \citep{monroe_fightin_2008}.

\begin{table}[t]
\centering
\small
\begin{tabular}{@{}lp{0.68\columnwidth}@{}}
\toprule
Group & Pretrained (roman) / Chat (\textit{italic}) \\
\midrule
African Am. & jazz, civil, community, rights, activist, justice, basketball \\
 & \textit{resilience, inspiring, resilient, community, jazz, hope, vibrant} \\
\midrule
Asian Am. & chinese, korean, parents, generation, english, japanese, immigrant \\
 & \textit{heritage, cultural, tech, san, generation, modern, identity} \\
\midrule
Hispanic Am. & spanish, mexico, mexican, english, family, immigrant, salsa \\
 & \textit{vibrant, heritage, proud, maria, cultural, salsa, culture} \\
\bottomrule
\end{tabular}
\caption{\textbf{In chat models, a shared vocabulary of identity and heritage
sets minority subjects apart.} Words most over-represented in text about each
minority group relative to text about White subjects, in the pretrained
(roman) and chat (italic) versions. Names and place fragments
(\textit{maria}, \textit{san}) are kept as generated.}
\label{tab:fightin}
\end{table}

In the pretrained models, the words that set minority subjects apart are
concrete associations, many of them stereotypes (Table~\ref{tab:fightin}).
Text about African Americans over-uses \textit{jazz}, \textit{civil rights}
and \textit{basketball}. Text about Asian Americans over-uses
\textit{Chinese}, \textit{parents} and \textit{immigrant}, and text about
Hispanic Americans \textit{Spanish}, \textit{Mexico} and \textit{immigrant}.
In the chat versions, most of the distinguishing words for all three groups
come from one shared vocabulary of identity and heritage, much of it positive
in tone: \textit{vibrant}, \textit{heritage}, \textit{cultural},
\textit{resilience}, \textit{identity}. A few group-specific words remain, such
as \textit{jazz} and \textit{salsa}.

\begin{figure}[t]
\centering
\includegraphics[width=\columnwidth]{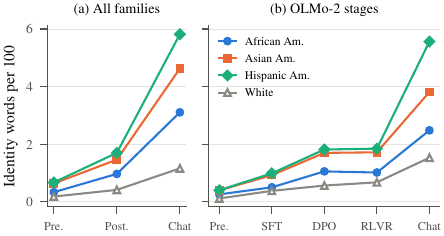}
\caption{\textbf{On average, identity words become more frequent after
post-training and common in chat use, mostly for minority subjects.} Identity
words per 100 words. (a)
Average over the seven families. (b) OLMo-2 by training stage (SFT:
supervised fine-tuning; DPO: preference optimization; RLVR: final stage) and
in chat use.}
\label{fig:vocab}
\end{figure}

A fixed list of 20 identity words (Appendix~\ref{app:lexicon},
Table~\ref{tab:vocab}) measures how
much of this vocabulary each version uses. Such words are rare in the
pretrained models (Figure~\ref{fig:vocab}): \idBaseHisp{} per 100 words for
Hispanic subjects and \idBaseWhite{} for White subjects. On average they become
more frequent after post-training and common in chat use, at \idChatHisp{} per 100
words for Hispanic, \idChatAsn{} for Asian and \idChatAfr{} for African
American subjects, against \idChatWhite{} for White subjects. The gap between
minority and White subjects grows in every family.

The similarity score does not register this vocabulary. Removing the identity
words leaves the homogeneity results in place, and text about African
Americans in chat models uses these words heavily yet is scored as no more
uniform than text about White Americans. In Llama-3.1, for example, text about
African Americans uses identity words \llamaChatAfrRatio{} times as often as
text about White Americans, yet its homogeneity effect is \llamaChatAfr~SD, or
\nmNmllamaChatAfr~SD with names hidden.

\section{Discussion}

When the request names the group, only two homogeneity biases hold across
the language models we tested. Text
about Hispanic Americans and about women is more uniform in all seven chat
models at the reference analysis, and neither effect was significantly
negative in more than \mvMaxSigNegHisp\% of any model's specifications. Both
effects remain when first names are hidden, and both are already present in
the pretrained models. The gender effect is the weaker of the two. It is
significant in four of the seven chat models, its prediction interval for a new
model reaches zero when first names are hidden (\nmNmPIWom), and the original
GPT-3.5 gender effect is not significant under the corrected test. Whether text about African
and Asian Americans is more or less uniform depends on the model. For African
Americans, these differences appear when the models are run as chat
assistants, and they disappear when the first names in the texts are hidden.
For Asian Americans, hiding names reduces the differences but does not remove
them.

\paragraph{A homogeneity score partly measures naming.} Chat models name most
of the people they write about, and how often they reuse the same names
differs from group to group and from model to model. A similarity
score computed on the full text mixes this naming pattern with the content of
the description. Audits that measure homogeneity should therefore report the
score with names hidden as well as without.

\paragraph{A low homogeneity score is not evidence of fair portrayal.} Much of
the identity vocabulary that chat models apply to minority subjects is positive
in tone, so filters for toxic or negative language are not designed to flag
it. It nevertheless expresses positive stereotypes
\citep{czopp_positive_2015}, it is applied far more to minority subjects than
to White subjects, and the similarity score does not detect it. An instruction
to portray each person as an individual reduces it by \indivDrop\% but does
not remove it (Appendix~\ref{app:indiv}).

\paragraph{Implications.} For auditors, when the request names the group
directly, a homogeneity finding about African or Asian Americans in one chat
model describes that model. Audits should
therefore test a model in the format in which it is deployed and report the
system message. Where a developer releases the pretrained model, testing it as
well shows which patterns are already present before post-training. Audit
frameworks \citep{mokander_auditing_2024} and the EU AI Act
\citep{eu_ai_act_2024} distinguish general-purpose models from the systems
built on them, and the chat format, which is supplied with the model but
applied when it is used, sits on that boundary. For researchers who use chat
models to write study materials or to simulate respondents, the results argue
for reporting the model and its system message, for checking key results with
a second model, and for expecting the identity vocabulary in text about
minority groups. For anyone measuring homogeneity bias, the pairwise mixed
model should be replaced by a test with one observation per prompt cell.

\section*{Limitations}

We study seven open model families of 7 to 20 billion parameters, English
text, and U.S.\ race categories. Proprietary
models do not release pretrained checkpoints, so the comparison across
training stages cannot be applied to them. The families differ in many ways
besides post-training, and we cannot say which ingredient of each recipe
produces each family's change. A pretrained model needs a different prompt
from a chat model. The prompt with three example texts holds the prompt fixed
between pretrained and post-trained weights, but the examples carry content
of their own, and the unnamed people in them may be read as White by
default. The chat version
differs from the post-trained version in prompt format only, so the chat
results describe each post-trained model under its developer's format; we did
not give the chat assistants the three example texts. The comparison between
versions rests on seven families. With surnames instead of group labels, the
chat models' African American estimates all have the same sign, so our
conclusions about African Americans concern requests that name the group
directly. The names analysis was run at one decoding setting. We
inherit four U.S.\ census race categories and binary gender from prior work.
Names are hidden with an automatic tagger, which misses some names and
occasionally tags other words, and every hidden name becomes the same
placeholder. Texts about Hispanic subjects name the person more often, so the
placeholder may itself add to the Hispanic effect with names hidden. The
cell-level test is conservative, so the null results reported here are weak
evidence of no effect. The main results rest on sentence encoders. A measure
of shared words agrees on the Hispanic effect but not on the gender effect,
which is positive in only three of seven models under it. No measure here has been validated against
human judgments of variety, so whether readers perceive these differences is
untested.

\section*{Ethics Statement}

This work measures a representational harm, the portrayal of some groups with
less variety than others. Generating that text reproduces stereotyped content,
some of which appears in the paper as word lists and examples. The race and
gender categories are social categories that are coarser than the identities
they label. We describe model outputs throughout and attribute no beliefs or
attitudes to the models. We release the generated texts and code at
\url{https://github.com/messihjlee/homogeneity-bias-robustness}.

\section*{Use of AI Assistants}

We used an AI coding assistant (Claude Code with Claude Opus 5.5, Anthropic) to
help write the generation and analysis code and to edit drafts of the paper.
The author designed the study, checked the code and its outputs, and verified
every claim. The language models studied here are research objects, not
assistants.

\appendix
\section{The Chat-model Multiverse}
\label{app:multiverse}

Section~\ref{sec:puzzle} summarizes this multiverse analysis of seven chat models.
A specification fixes the model, decoding setting (temperature
$\{0.5, 1, 1.5, 2\}$ $\times$ top-$p$ $\{.2, .4, .6, .8, 1\}$), prompt wording,
identity cue, text representation and cue treatment. Prompt wordings change one
element of the original at a time: no system message, the system message
\textit{``You are a helpful assistant.''}, or the group terms \textit{Black},
\textit{Latino/Latina} and \textit{White}. The surname cue replaces the group
label with one of 50 surnames per group whose bearers in the 2010 Census
\citep{comenetz_frequently_2016} are over 70\% in the target group and under
20\% in any other. Representations are seven sentence encoders
(three Sentence-BERT models, BGE, E5, GTE and GTR-T5) and a lexical Jaccard
measure. Each specification is crossed with seven models, eight measures and
two cue treatments (raw, masked), which gives 5{,}120 in all: 2{,}240 for the
original wording over the 20 decoding settings, 112 for each of the three
alternative wordings and 192 for the top-$k$-off arm at one or two settings,
and 2{,}352 for the surname cue, which adds temperature 0.

\begin{table*}[t]
\centering
\small
\begin{tabular}{@{}llrcrrr@{}}
\toprule
 & & \multicolumn{3}{c}{Reference spec.} & \multicolumn{2}{c}{Multiverse} \\
\cmidrule(lr){3-5}\cmidrule(l){6-7}
Cue & Contrast & Pooled & 95\% PI & $>0$ & Sig.$+$ & Sig.$-$ \\
\midrule
Label & African Am. & +0.05 & [$-$0.43, +0.53] & 4/7 & 8\% & 9\% \\
 & African Am. (m) & $-$0.05 & [$-$0.49, +0.40] & 3/7 & 6\% & 13\% \\
 & Asian Am. & +0.12 & [$-$0.36, +0.60] & 5/7 & 20\% & 2\% \\
 & Asian Am. (m) & $-$0.06 & [$-$0.47, +0.35] & 3/7 & 8\% & 7\% \\
 & Hispanic Am. & +0.53 & [+0.23, +0.82] & 7/7 & 51\% & 0\% \\
 & Hispanic Am. (m) & +0.35 & [+0.10, +0.60] & 7/7 & 46\% & 0\% \\
 & Woman & +0.13 & [+0.03, +0.22] & 7/7 & 17\% & 0\% \\
 & Woman (m) & +0.16 & [+0.08, +0.25] & 7/7 & 22\% & 0\% \\
\midrule
Surname & Black & $-$0.31 & [$-$0.39, $-$0.22] & 0/7 & 0\% & 66\% \\
 & Black (m) & $-$0.10 & [$-$0.16, $-$0.03] & 1/7 & 0\% & 45\% \\
 & Asian Am. & +0.19 & [+0.12, +0.26] & 7/7 & 49\% & 0\% \\
 & Asian Am. (m) & +0.16 & [+0.06, +0.25] & 7/7 & 17\% & 0\% \\
 & Hispanic Am. & +0.37 & [+0.22, +0.51] & 7/7 & 60\% & 0\% \\
 & Hispanic Am. (m) & +0.03 & [$-$0.04, +0.09] & 5/7 & 0\% & 1\% \\
\bottomrule
\end{tabular}
\caption{Chat-model multiverse. \textit{Reference spec.}: pooled random-effects
estimate across seven chat models at temperature 1, top-$p$ 1 with
\texttt{all-mpnet-base-v2}, its 95\% prediction interval for a new model, and the
number of models with a positive estimate. \textit{Multiverse}: share of all
specifications for that cue and treatment that are significantly positive or
negative after Benjamini--Hochberg adjustment. (m): group term or surname
masked.}
\label{tab:summary}
\end{table*}

\begin{figure*}[t]
\centering
\includegraphics[width=\textwidth]{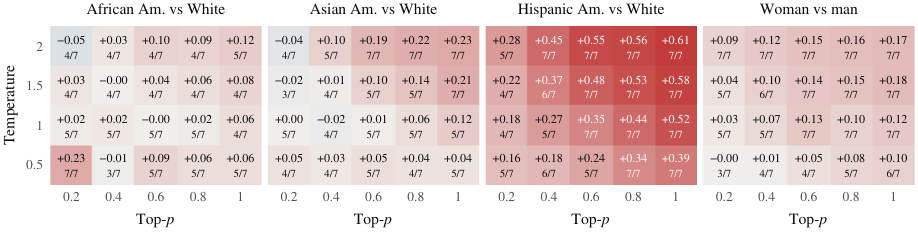}
\caption{Chat models: mean standardized contrast across the seven models at each
decoding setting (raw text, \texttt{all-mpnet-base-v2}). Small numbers: models
with a positive estimate.}
\label{fig:decoding}
\end{figure*}

\paragraph{Findings.} Decoding rescales effects almost fourfold, from $+0.16$
(temperature 0.5, top-$p$ 0.2) to $+0.61$ (temperature 2, top-$p$ 1) for the
Hispanic contrast on unmasked text (Figure~\ref{fig:decoding}). Signs are less
stable where sampling is narrow (low top-$p$ or temperature 0.5). With the group
term masked, the Hispanic contrast is positive in all seven models in 9 of the
20 decoding settings and in at least four models in every setting. The gender
contrast is positive in all seven models in 12 settings and in at least three
in every setting. Of the 140 model-by-setting estimates for each contrast, two
Hispanic estimates and no gender estimates are significantly negative
($p<.05$, unadjusted).
The Hispanic and gender contrasts keep their direction under every wording; the
pooled African American contrast changes sign between wordings. The Asian
American contrast is positive in all seven models under BGE and GTE and split
under the other encoders. With surnames as the cue, the Hispanic effect is
carried by the surnames themselves: the 50 Hispanic surnames are more similar to
one another than any other group's before any model writes about them, and
masking the surname removes the effect. A variance decomposition of the
2{,}240 label-cue estimates per contrast (analysis of variance with all two-way
interactions) gives the shares in Table~\ref{tab:variance}. The model is the
largest single factor for every contrast.

\begin{table}[h]
\centering
\small
\resizebox{\columnwidth}{!}{\begin{tabular}{@{}lrrrr@{}}
\toprule
Source & African & Asian & Hispanic & Woman \\
\midrule
Model & 52 & 26 & 40 & 26 \\
Decoding & 10 & 13 & 14 & 24 \\
Model $\times$ decoding & 17 & 19 & 15 & 12 \\
Measure & 2 & 12 & 9 & 10 \\
Measure $\times$ others & 6 & 8 & 7 & 8 \\
Masking & 0 & 5 & 2 & 0 \\
Masking $\times$ others & 0 & 2 & 0 & 1 \\
Residual & 13 & 15 & 12 & 19 \\
\bottomrule
\end{tabular}}
\caption{Share of variance (\%) in the label-cue estimates. \textit{Decoding}:
temperature, top-$p$ and their interaction. \textit{Measure}: seven sentence
encoders and Jaccard similarity. \textit{Residual}: interactions of three or
more factors.}
\label{tab:variance}
\end{table}

\paragraph{Refusals.} gpt-oss refuses 37\% of prompts about White Americans and
under 4\% of prompts about other groups. A refusal paired with a story has near-zero
similarity, so refusals left in a cell lower its mean pairwise cosine
similarity (MPCS) and inflate every minority
contrast for that model. All analyses remove refusals before measurement.

\section{Models}
\label{app:models}

\begin{table}[h]
\centering
\small
\resizebox{\columnwidth}{!}{\begin{tabular}{@{}lll@{}}
\toprule
Family & Pretrained & Post-trained \\
\midrule
Llama-3.1-8B & \texttt{Llama-3.1-8B} & \texttt{-Instruct} \\
Qwen3-8B & \texttt{Qwen3-8B-Base} & \texttt{Qwen3-8B} \\
Falcon3-10B & \texttt{Falcon3-10B-Base} & \texttt{-Instruct} \\
Mistral-NeMo-12B & \texttt{Mistral-Nemo-Base-2407} & \texttt{-Instruct-2407} \\
OLMo-2-7B & \texttt{OLMo-2-1124-7B} & \texttt{-SFT}, \texttt{-DPO}, \texttt{-Instruct} \\
Granite-3.3-8B & \texttt{granite-3.3-8b-base} & \texttt{-instruct} \\
Gemma-2-9B & \texttt{gemma-2-9b} & \texttt{-it} \\
gpt-oss-20b & --- & \texttt{gpt-oss-20b} \\
\bottomrule
\end{tabular}}
\caption{Checkpoints (Hugging Face), run in bf16 with \texttt{transformers}:
Llama-3.1 \citep{dubey_llama_2024}, Qwen3 \citep{qwen3_2025}, Falcon3
\citep{falcon3_2024}, Mistral-NeMo \citep{team_mistral_2024}, OLMo-2
\citep{olmo_2_2025}, Granite-3.3 \citep{granite3_2025}, Gemma-2
\citep{gemma2_2024}, and gpt-oss \citep{gptoss_2025}.
OLMo-2's final \texttt{Instruct} checkpoint follows DPO with reinforcement
learning from verifiable rewards (RLVR). gpt-oss appears only in the chat-model
analyses.}
\label{tab:models}
\end{table}

\paragraph{Decoding.} Temperature 1, top-$p$ 1 (and 0.8 for the replication),
top-$k$ at the library default of 50, except Qwen3, whose post-trained
checkpoint ships $k=20$; its pretrained checkpoint was run at $k=20$ as well
(a run at $k=50$ gives nearly the same estimates: $+0.09$, $+0.13$, $+0.30$
and $+0.24$ SD for the African, Asian, Hispanic and gender contrasts, against
$+0.12$, $+0.14$, $+0.30$ and $+0.24$ at $k=20$). At most 80
new tokens. In the chat conditions, Qwen3 runs with thinking disabled, and
gpt-oss at low reasoning effort with only its final channel kept. Gemma-2's
template has no system role, so the system text is prepended to the user turn.

\section{The Few-shot Frame}
\label{app:frame}

Every pretrained and post-trained checkpoint in the few-shot conditions receives
the following text, with the last line varying by cell:

\begin{small}
\begin{quote}\raggedright
\texttt{Text 1: a thirty-word story about a firefighter.}\\
\texttt{After the night shift, the firefighter drove to the lake instead of home.
She skipped stones until sunrise, counting each one, then finally slept in the
car with the windows down.}\\[3pt]
\texttt{Text 2: a thirty-word biography of a retired teacher.}\\
\texttt{Born in a small mill town, he taught chemistry for thirty-four years,
coached the chess club, restored two vintage motorcycles, and now volunteers at
the county library every Thursday afternoon.}\\[3pt]
\texttt{Text 3: a thirty-word social media profile of a college student.}\\
\texttt{Second-year civil engineering major. Night-owl, bad guitarist, decent
cook. Currently building a bridge out of popsicle sticks for class. Ask me about
bus routes, I know all of them.}\\[3pt]
\texttt{Text 4: a thirty-word \{format\} a(n) \{race\} American \{gender\}.}\\
\textit{(\{format\} carries its preposition, e.g.\ ``story about'', ``biography of''.)}
\end{quote}
\end{small}

\noindent The continuation is cut at the first blank line or at the next
\texttt{Text $n$:} header. The exemplars state no race or ethnicity; one uses a
female and one a male pronoun. A zero-shot frame
(\textit{``The following is a thirty-word story about \dots''}) was tried first on
Llama-3.1 and abandoned because the pretrained model mostly produced headings and
meta-text; no analysis was run on it.

\paragraph{Examples.} One cell, \textit{``story about a(n) Hispanic American
woman''}, first sampled text in each condition (Llama-3.1-8B):
\begin{small}
\begin{itemize}\itemsep1pt
\item \textbf{Pretrained.} ``The only daughter of immigrants, she started her own business when she was twenty-two. She is fluent in English and Spanish, and likes to cook tamales when she has time. She is considering running for councilwoman.''
\item \textbf{Post-trained.} ``As the sun set over the Rio Grande, she began to salsa dance in the empty street, her shoulders swinging, her hips swaying, the music echoing off the worn adobe walls of her village.''
\item \textbf{Chat.} ``Luz, a vibrant Hispanic American artist, wove together her Puerto Rican heritage and Texas roots in her paintings, infusing vibrant hues and eclectic styles that captured the beauty of her vibrant cultural identity.''
\item \textbf{Chat + instr.} ``As she sat on the beach, Maria sipped her café con leche and gazed out at the Pacific, her bright smile betraying a deep sense of peace after a long week of work.''
\end{itemize}
\end{small}

\section{Output Quality and Global Homogeneity}
\label{app:qc}

\begin{table}[h]
\centering
\small
\resizebox{\columnwidth}{!}{\begin{tabular}{@{}llrrrr@{}}
\toprule
Family & Condition & Kept \% & Words & MPCS & Term \% \\
\midrule
Llama-3.1-8B & Pretrained & 99.8 & 34 & 0.34 & 19 \\
 & Post-trained & 100.0 & 32 & 0.41 & 5 \\
 & Chat & 100.0 & 31 & 0.57 & 44 \\
 & Chat + instr. & 100.0 & 32 & 0.52 & 21 \\
Qwen3-8B & Pretrained & 100.0 & 35 & 0.40 & 22 \\
 & Post-trained & 100.0 & 33 & 0.43 & 22 \\
 & Chat & 99.7 & 39 & 0.74 & 68 \\
 & Chat + instr. & 100.0 & 41 & 0.70 & 47 \\
Falcon3-10B & Pretrained & 98.4 & 31 & 0.39 & 18 \\
 & Post-trained & 99.9 & 28 & 0.37 & 41 \\
 & Chat & 99.2 & 24 & 0.63 & 87 \\
 & Chat + instr. & 99.6 & 27 & 0.57 & 79 \\
Mistral-NeMo-12B & Pretrained & 99.8 & 36 & 0.32 & 10 \\
 & Post-trained & 100.0 & 28 & 0.43 & 7 \\
 & Chat & 99.9 & 23 & 0.54 & 44 \\
 & Chat + instr. & 100.0 & 31 & 0.52 & 47 \\
OLMo-2-7B & Pretrained & 100.0 & 36 & 0.35 & 11 \\
 & SFT & 100.0 & 34 & 0.41 & 15 \\
 & DPO & 100.0 & 33 & 0.46 & 14 \\
 & Post-trained & 100.0 & 31 & 0.46 & 13 \\
 & Chat & 100.0 & 39 & 0.58 & 69 \\
 & Chat + instr. & 100.0 & 32 & 0.48 & 21 \\
Granite-3.3-8B & Pretrained & 97.5 & 34 & 0.30 & 9 \\
 & Post-trained & 99.9 & 30 & 0.46 & 12 \\
 & Chat & 100.0 & 30 & 0.58 & 65 \\
 & Chat + instr. & 100.0 & 34 & 0.55 & 38 \\
Gemma-2-9B & Pretrained & 99.9 & 40 & 0.35 & 13 \\
 & Post-trained & 99.7 & 25 & 0.44 & 3 \\
 & Chat & 98.2 & 25 & 0.51 & 14 \\
 & Chat + instr. & 100.0 & 23 & 0.48 & 2 \\
gpt-oss-20b & Chat & 89.2 & 29 & 0.58 & 51 \\
\bottomrule
\end{tabular}}
\caption{Share of texts kept after removing refusals and outputs under eight
words; median words per text; mean within-cell MPCS (raw text); and the share of
texts that restate the group term (\textit{Term}). Temperature 1, top-$p$ 1.}
\label{tab:qc}
\end{table}

Output length barely varies by group within a model and version: mean words
per text differ across the eight race $\times$ gender groups by a median of
\lenRelMed\% (at most \lenRelMax\%, or \lenWordsMax{} words). Since contrasts
compare groups within a model and version, length differences between models
do not enter them.

Post-training raises within-cell similarity in every family but Falcon3, and
the chat condition raises it further in all of them (Table~\ref{tab:qc}). It also makes the model restate
the group term far more often, which is why the primary estimates mask it.

\section{Robustness}
\label{app:robust}

Table~\ref{tab:divergence} repeats the pooled analysis of
Sections~\ref{sec:agree}--\ref{sec:content} under five variants: raw text instead of
masked; estimates in cosine units rather than SD units, which removes any effect
of the condition's overall spread on the scale; the identity lexicon masked as
well as the group term; the BGE encoder, under which the chat models' Asian
American contrast differed most from \texttt{all-mpnet-base-v2} in the
multiverse; and top-$p$ 0.8.

\begin{table*}[t]
\centering
\small
\begin{tabular}{@{}lrrrr@{}}
\toprule
Condition & African Am. & Asian Am. & Hispanic Am. & Woman \\
\midrule
\multicolumn{5}{@{}l}{\textit{Main (masked, SD units)}} \\
\quad Pretrained ($k$=7) & $+0.04$ (0.05) & $+0.04$ (0.05) & $+0.24$ (0.07$^{\dagger}$) & $+0.22$ (0.00) \\
\quad Post-trained ($k$=7) & $+0.01$ (0.00) & $+0.12$ (0.08$^{\dagger}$) & $+0.39$ (0.08$^{\dagger}$) & $+0.19$ (0.04) \\
\quad Chat ($k$=7) & $-0.08$ (0.13$^{\dagger}$) & $-0.04$ (0.15$^{\dagger}$) & $+0.38$ (0.11$^{\dagger}$) & $+0.17$ (0.03) \\
\quad Chat + instr. ($k$=7) & $-0.19$ (0.16$^{\dagger}$) & $-0.16$ (0.10$^{\dagger}$) & $+0.39$ (0.24$^{\dagger}$) & $+0.17$ (0.05) \\
\multicolumn{5}{@{}l}{\textit{Raw text}} \\
\quad Pretrained ($k$=7) & $+0.06$ (0.07$^{\dagger}$) & $+0.11$ (0.08$^{\dagger}$) & $+0.28$ (0.10$^{\dagger}$) & $+0.19$ (0.00) \\
\quad Post-trained ($k$=7) & $+0.05$ (0.03) & $+0.19$ (0.13$^{\dagger}$) & $+0.45$ (0.08$^{\dagger}$) & $+0.16$ (0.05$^{\dagger}$) \\
\quad Chat ($k$=7) & $+0.01$ (0.16$^{\dagger}$) & $+0.13$ (0.17$^{\dagger}$) & $+0.55$ (0.11$^{\dagger}$) & $+0.14$ (0.04) \\
\quad Chat + instr. ($k$=7) & $-0.11$ (0.19$^{\dagger}$) & $-0.03$ (0.15$^{\dagger}$) & $+0.52$ (0.23$^{\dagger}$) & $+0.15$ (0.05) \\
\multicolumn{5}{@{}l}{\textit{Cosine units}} \\
\quad Pretrained ($k$=7) & $+0.005$ (0.007) & $+0.006$ (0.006) & $+0.033$ (0.010$^{\dagger}$) & $+0.030$ (0.000) \\
\quad Post-trained ($k$=7) & $+0.001$ (0.000) & $+0.017$ (0.012$^{\dagger}$) & $+0.054$ (0.012$^{\dagger}$) & $+0.026$ (0.004) \\
\quad Chat ($k$=7) & $-0.013$ (0.020$^{\dagger}$) & $-0.005$ (0.023$^{\dagger}$) & $+0.058$ (0.018$^{\dagger}$) & $+0.026$ (0.005) \\
\quad Chat + instr. ($k$=7) & $-0.028$ (0.023$^{\dagger}$) & $-0.023$ (0.016$^{\dagger}$) & $+0.058$ (0.036$^{\dagger}$) & $+0.025$ (0.008) \\
\multicolumn{5}{@{}l}{\textit{Identity lexicon masked}} \\
\quad Pretrained ($k$=7) & $+0.03$ (0.05) & $+0.04$ (0.05$^{\dagger}$) & $+0.23$ (0.07$^{\dagger}$) & $+0.22$ (0.00) \\
\quad Post-trained ($k$=7) & $+0.01$ (0.00) & $+0.12$ (0.09$^{\dagger}$) & $+0.38$ (0.10$^{\dagger}$) & $+0.19$ (0.03) \\
\quad Chat ($k$=7) & $-0.07$ (0.11$^{\dagger}$) & $+0.01$ (0.11$^{\dagger}$) & $+0.38$ (0.09$^{\dagger}$) & $+0.16$ (0.04) \\
\quad Chat + instr. ($k$=7) & $-0.17$ (0.15$^{\dagger}$) & $-0.11$ (0.09$^{\dagger}$) & $+0.40$ (0.20$^{\dagger}$) & $+0.16$ (0.05) \\
\multicolumn{5}{@{}l}{\textit{BGE encoder}} \\
\quad Pretrained ($k$=7) & $-0.04$ (0.05) & $+0.11$ (0.08$^{\dagger}$) & $+0.19$ (0.04) & $+0.09$ (0.00) \\
\quad Post-trained ($k$=7) & $+0.01$ (0.08$^{\dagger}$) & $+0.23$ (0.08$^{\dagger}$) & $+0.36$ (0.11$^{\dagger}$) & $+0.08$ (0.00) \\
\quad Chat ($k$=7) & $-0.02$ (0.08) & $+0.13$ (0.08$^{\dagger}$) & $+0.34$ (0.00) & $+0.08$ (0.02) \\
\quad Chat + instr. ($k$=7) & $-0.08$ (0.14$^{\dagger}$) & $+0.06$ (0.10$^{\dagger}$) & $+0.33$ (0.14$^{\dagger}$) & $+0.07$ (0.06$^{\dagger}$) \\
\multicolumn{5}{@{}l}{\textit{Top-$p$ 0.8}} \\
\quad Pretrained ($k$=7) & $+0.02$ (0.05) & $+0.04$ (0.06$^{\dagger}$) & $+0.20$ (0.08$^{\dagger}$) & $+0.20$ (0.00) \\
\quad Post-trained ($k$=7) & $-0.01$ (0.04) & $+0.08$ (0.10$^{\dagger}$) & $+0.33$ (0.10$^{\dagger}$) & $+0.17$ (0.05$^{\dagger}$) \\
\quad Chat ($k$=7) & $-0.12$ (0.23$^{\dagger}$) & $-0.09$ (0.15$^{\dagger}$) & $+0.29$ (0.18$^{\dagger}$) & $+0.14$ (0.00) \\
\bottomrule
\end{tabular}
\caption{Pooled random-effects estimates across the families that have a
pretrained checkpoint, with between-family $\tau$ in parentheses.
$^{\dagger}$Cochran's $Q$ $p<.05$ (significant heterogeneity between families).
SD units except \textit{Cosine units}.}
\label{tab:divergence}
\end{table*}

\paragraph{Where the split holds.} The comparison that carries the argument is
the between-family spread $\tau$ of the African and Asian American contrasts
in chat relative to the pretrained models. Computed from unrounded $\tau$, it is 2.4 and 3.3 times larger in
the main analysis, 2.3--2.4 with raw text, 2.7--3.6 in cosine units, 2.0--2.1
with the identity lexicon masked and 2.6--4.2 at top-$p$ 0.8, and the change
from the pretrained model differs significantly between families in each of
these except the Asian American contrast with the lexicon masked ($p=.087$).
\paragraph{Seven encoders.} Table~\ref{tab:encoders} repeats the main
analysis under all seven sentence encoders of the multiverse. For African
Americans the result holds under \encChatSigAfr{} of \encN: the pretrained
spread is at most \encTauBaseMaxAfr, the chat spread \encTauChatRangeAfr, and
the change to chat differs between families, while the change from
post-training alone does so under \encInstSigAfr{} of \encN. The exception is
BGE, where $\tau$ rises only from 0.05 to 0.08. For Asian Americans the chat
spread is larger than the pretrained spread under every encoder except BGE,
but the change test is significant under only \encChatSigAsn{} of \encN; we
treat the Asian American split as suggestive. Under every encoder,
post-training alone changes the Hispanic contrast by family-specific amounts
(\encInstSigHisp{} of \encN) without changing its sign, and the gender
contrast never splits. Post-training alone moves the Asian American contrast
by family-specific amounts under one encoder (BGE). The last row repeats the
analysis with Jaccard similarity of word sets, the lexical measure of the
multiverse. It finds no pretrained Hispanic contrast (\jacBaseHisp) and no
split in chat ($p_{\Delta} = \jacChatDeltaQpAfr$ and \jacChatDeltaQpAsn),
though it does show family-specific change after post-training alone. The
contrasts this paper reports concern meaning, not shared vocabulary.

\begin{table*}[t]
\centering
\small
\begin{tabular}{@{}lrrrrrrrr@{}}
\toprule
 & \multicolumn{4}{c}{African Am.} & \multicolumn{4}{c}{Asian Am.} \\
\cmidrule(lr){2-5}\cmidrule(l){6-9}
Encoder & $\tau_{\text{pre}}$ & $\tau_{\text{chat}}$ & $p_{\Delta}^{\text{post}}$ & $p_{\Delta}^{\text{chat}}$ & $\tau_{\text{pre}}$ & $\tau_{\text{chat}}$ & $p_{\Delta}^{\text{post}}$ & $p_{\Delta}^{\text{chat}}$ \\
\midrule
all-mpnet-base-v2 (main) & 0.05 & 0.13 & .482 & \textbf{.007} & 0.05 & 0.15 & .709 & \textbf{.009} \\
all-MiniLM-L12-v2 & 0.04 & 0.14 & .473 & \textbf{.003} & 0.05 & 0.13 & .367 & .074 \\
all-distilroberta-v1 & 0.05 & 0.13 & .482 & \textbf{.002} & 0.04 & 0.13 & .348 & \textbf{.010} \\
bge-base-en-v1.5 & 0.05 & 0.08 & .162 & .227 & 0.08 & 0.08 & \textbf{.040} & .590 \\
e5-base-v2 & 0.00 & 0.15 & .491 & \textbf{.006} & 0.06 & 0.14 & .370 & .257 \\
gte-base & 0.00 & 0.12 & .735 & \textbf{.029} & 0.06 & 0.12 & .561 & .095 \\
gtr-t5-base & 0.02 & 0.15 & .237 & \textbf{.001} & 0.06 & 0.12 & .853 & .139 \\
\midrule
Jaccard (word sets) & 0.03 & 0.06 & \textbf{.047} & .270 & 0.05 & 0.06 & \textbf{.003} & .313 \\
\bottomrule
\end{tabular}
\caption{Pretrained versus chat under each sentence encoder (group term masked,
SD units, seven families). $\tau$: between-family SD. $p_{\Delta}$: Cochran's
$Q$ on each family's change from the pretrained model, for the post-trained
model with the identical prompt (\textit{post}) and for chat use
(\textit{chat}); bold $p<.05$. Last row: Jaccard similarity of word sets
instead of an encoder.}
\label{tab:encoders}
\end{table*}

\paragraph{Other variants.} The pretrained models' Asian American
spread is itself significant at top-$p$ 0.8 ($\tau=0.06$, $p=.047$) and with
the identity lexicon masked ($\tau=0.05$, $p=.049$), as both pretrained spreads
are with raw text; under BGE the pretrained Asian American spread is
significant ($p=.001$) and the chat spread no larger.

\paragraph{Masking the identity register.} Masking the identity lexicon as well
as the group term leaves the chat models' Hispanic contrast where it was
(\chatHisp{} $\to$ \chatHispId; pretrained \baseHisp{} $\to$ \baseHispId) and
changes no conclusion of Section~\ref{sec:split}. The register and the
homogeneity contrasts therefore dissociate: in chat, the African and Asian
American homogeneity contrasts are near zero or negative, yet identity words
are \idChatAfr{} and \idChatAsn{} per 100 for those groups against
\idChatWhite{} for White subjects. In Llama-3.1, text about African Americans is
\emph{more} varied than text about White Americans (\llamaChatAfr) while using
identity words \llamaChatAfrRatio{} times as often. Masking these words leaves the homogeneity estimates in place, so an audit
by embedding homogeneity alone would not register this change.

\paragraph{Global compression.} Post-training raises within-cell similarity on
average (raw text): mean MPCS is \mpcsBase{} in the pretrained condition, \mpcsInst{}
with post-trained weights and \mpcsChat{} in chat (Table~\ref{tab:qc}). Because
the main contrasts are scaled by each condition's spread, this global narrowing
is not what Figure~\ref{fig:arms} shows; the cosine-unit row of
Table~\ref{tab:divergence} leads to the same conclusions.

\clearpage
\section{Per-family Estimates}
\label{app:families}

\begin{table}[!ht]
\centering
\small
\resizebox{\columnwidth}{!}{\begin{tabular}{@{}llrrrr@{}}
\toprule
Family & Condition & African Am. & Asian Am. & Hispanic Am. & Woman \\
\midrule
Llama-3.1-8B & Pretrained & $+0.14$$^{*}$ & $+0.07$ & $+0.29$$^{*}$ & $+0.18$$^{*}$ \\
 & Post-trained & $-0.03$ & $+0.16$$^{*}$ & $+0.43$$^{*}$ & $+0.30$$^{*}$ \\
 & Chat & $-0.29$$^{*}$ & $-0.18$$^{*}$ & $+0.30$$^{*}$ & $+0.08$ \\
 & Chat + instr. & $-0.50$$^{*}$ & $-0.40$$^{*}$ & $+0.06$ & $+0.12$$^{*}$ \\
\addlinespace[2pt]
Qwen3-8B & Pretrained & $+0.12$$^{*}$ & $+0.14$$^{*}$ & $+0.30$$^{*}$ & $+0.24$$^{*}$ \\
 & Post-trained & $+0.06$ & $+0.11$ & $+0.26$$^{*}$ & $+0.20$$^{*}$ \\
 & Chat & $+0.20$ & $+0.06$ & $+0.29$$^{*}$ & $+0.16$ \\
 & Chat + instr. & $+0.23$ & $-0.15$ & $+0.28$$^{*}$ & $+0.22$$^{*}$ \\
\addlinespace[2pt]
Falcon3-10B & Pretrained & $+0.01$ & $+0.11$ & $+0.41$$^{*}$ & $+0.22$$^{*}$ \\
 & Post-trained & $-0.01$ & $+0.23$$^{*}$ & $+0.27$$^{*}$ & $+0.15$$^{*}$ \\
 & Chat & $+0.09$ & $+0.19$$^{*}$ & $+0.17$$^{*}$ & $+0.18$$^{*}$ \\
 & Chat + instr. & $-0.14$ & $-0.10$ & $+0.15$$^{*}$ & $+0.20$$^{*}$ \\
\addlinespace[2pt]
Mistral-NeMo-12B & Pretrained & $+0.03$ & $+0.01$ & $+0.22$$^{*}$ & $+0.26$$^{*}$ \\
 & Post-trained & $-0.06$ & $+0.12$ & $+0.32$$^{*}$ & $+0.21$$^{*}$ \\
 & Chat & $-0.01$ & $+0.04$ & $+0.46$$^{*}$ & $+0.21$$^{*}$ \\
 & Chat + instr. & $-0.16$$^{*}$ & $-0.15$$^{*}$ & $+0.64$$^{*}$ & $+0.21$$^{*}$ \\
\addlinespace[2pt]
OLMo-2-7B & Pretrained & $-0.10$ & $-0.10$ & $+0.08$ & $+0.19$$^{*}$ \\
 & Post-trained & $-0.05$ & $-0.10$ & $+0.46$$^{*}$ & $+0.12$$^{*}$ \\
 & Chat & $-0.15$$^{*}$ & $-0.13$$^{*}$ & $+0.41$$^{*}$ & $+0.23$$^{*}$ \\
 & Chat + instr. & $-0.25$$^{*}$ & $-0.23$$^{*}$ & $+0.50$$^{*}$ & $+0.04$ \\
\addlinespace[2pt]
Granite-3.3-8B & Pretrained & $+0.01$ & $+0.04$ & $+0.21$$^{*}$ & $+0.24$$^{*}$ \\
 & Post-trained & $+0.09$ & $+0.20$$^{*}$ & $+0.49$$^{*}$ & $+0.17$$^{*}$ \\
 & Chat & $-0.26$$^{*}$ & $-0.27$$^{*}$ & $+0.45$$^{*}$ & $+0.09$ \\
 & Chat + instr. & $-0.23$$^{*}$ & $-0.05$ & $+0.63$$^{*}$ & $+0.17$$^{*}$ \\
\addlinespace[2pt]
Gemma-2-9B & Pretrained & $+0.03$ & $+0.02$ & $+0.22$$^{*}$ & $+0.19$$^{*}$ \\
 & Post-trained & $+0.06$ & $+0.13$$^{*}$ & $+0.52$$^{*}$ & $+0.15$$^{*}$ \\
 & Chat & $-0.06$ & $+0.08$ & $+0.61$$^{*}$ & $+0.24$$^{*}$ \\
 & Chat + instr. & $-0.08$ & $-0.01$ & $+0.45$$^{*}$ & $+0.26$$^{*}$ \\
\addlinespace[2pt]
gpt-oss-20b & Chat & $+0.19$ & $-0.11$ & $+0.30$$^{*}$ & $+0.19$$^{*}$ \\
\bottomrule
\end{tabular}}
\caption{Contrasts by family and condition, group term masked, SD units.
$^{*}|z|>1.96$.}
\label{tab:families}
\end{table}

\paragraph{OLMo-2 stage by stage.} OLMo-2 also releases the checkpoints after
supervised fine-tuning (SFT) and preference optimization (DPO) that precede its
final reinforcement-learning stage \citep{olmo_2_2025}, following the
T\"ulu~3 recipe \citep{lambert_tulu_2024}. Its Hispanic
contrast is weak in the pretrained checkpoint (\olmoBaseHisp), grows after SFT
(\olmoSftHisp) and DPO (\olmoDpoHisp), and stays there through the final stage
(\olmoInstHisp). The African and Asian American contrasts, slightly negative in
the pretrained checkpoint, stay so throughout and become significantly negative
only in chat (\olmoChatAfr, \olmoChatAsn; Table~\ref{tab:olmo}).

\begin{table}[!ht]
\centering
\small
\resizebox{\columnwidth}{!}{\begin{tabular}{@{}lrrrrr@{}}
\toprule
Checkpoint & African Am. & Asian Am. & Hispanic Am. & Woman & ID gap \\
\midrule
Pretrained & $-0.10$ & $-0.10$ & $+0.08$ & $+0.19$$^{*}$ & 0.24 \\
SFT & $-0.05$ & $-0.10$ & $+0.26$$^{*}$ & $+0.21$$^{*}$ & 0.43 \\
DPO & $-0.02$ & $-0.06$ & $+0.44$$^{*}$ & $+0.15$$^{*}$ & 0.96 \\
Post-trained & $-0.05$ & $-0.10$ & $+0.46$$^{*}$ & $+0.12$$^{*}$ & 0.85 \\
Chat & $-0.15$$^{*}$ & $-0.13$$^{*}$ & $+0.41$$^{*}$ & $+0.23$$^{*}$ & 2.42 \\
Chat + instr. & $-0.25$$^{*}$ & $-0.23$$^{*}$ & $+0.50$$^{*}$ & $+0.04$ & 1.23 \\
\bottomrule
\end{tabular}}
\caption{OLMo-2-7B by checkpoint (masked, SD units). \textit{ID gap}: identity
words per 100 for minority subjects minus White subjects. All but \textit{Chat}
use the few-shot prompt. $^{*}|z|>1.96$.}
\label{tab:olmo}
\end{table}

\section{Identity Lexicon}
\label{app:lexicon}

The lexicon matches, case-insensitively and as whole words:
\textit{heritage, culture, cultural, cultures, roots, proud, pride, vibrant,
resilien*, tradition*, ancest*, community, communities, immigra*, identity,
diaspora, legacy, homeland, bicultural, first-generation}. It was fixed after
inspecting the Llama-3.1 pilot (pretrained versus chat) and before any other
family was generated. The log-odds analysis in Table~\ref{tab:fightin} uses no
lexicon and recovers the same words.

\begin{table}[h]
\centering
\small
\resizebox{\columnwidth}{!}{\begin{tabular}{@{}lrrrr@{}}
\toprule
Family & Pretrained & Post-trained & Chat & Chat + instr. \\
\midrule
Llama-3.1-8B & 0.5 / 0.2 & 1.1 / 0.2 & 5.2 / 0.8 & 3.0 / 0.4 \\
Qwen3-8B & 1.0 / 0.3 & 1.4 / 0.5 & 7.0 / 1.7 & 4.8 / 1.0 \\
Falcon3-10B & 0.7 / 0.2 & 0.6 / 0.2 & 5.0 / 1.5 & 4.2 / 1.6 \\
Mistral-NeMo-12B & 0.4 / 0.2 & 1.0 / 0.4 & 3.3 / 0.5 & 1.9 / 0.4 \\
OLMo-2-7B & 0.3 / 0.1 & 1.5 / 0.7 & 4.0 / 1.5 & 2.0 / 0.8 \\
Granite-3.3-8B & 0.3 / 0.2 & 2.5 / 0.6 & 4.8 / 1.7 & 3.9 / 1.0 \\
Gemma-2-9B & 0.6 / 0.2 & 1.5 / 0.4 & 2.4 / 0.3 & 1.0 / 0.1 \\
gpt-oss-20b & -- & -- & 3.9 / 1.5 & -- \\
\bottomrule
\end{tabular}}
\caption{Identity words per 100 words: mean over the three minority groups /
White subjects, by family and condition.}
\label{tab:vocab}
\end{table}

\section{An Instruction to Individuate}
\label{app:indiv}

A fourth condition, \textit{Chat + instruction}, adds one sentence to the chat
system message: \textit{``Portray each person as an individual, not as a
representative of their racial or ethnic group.''} It lowers identity words for
minority subjects by \indivDrop\% on average but does not remove them: they
remain \indivRatio{} times as frequent as for White subjects. The homogeneity
contrasts move in different directions. The African and Asian American contrasts
fall below zero in most families (\indivAfr{} and \indivAsn): the instruction does not
remove these contrasts but reverses them, and text about these groups becomes
more varied than text about White subjects. The Hispanic contrast (\indivHisp) and the gender contrast
(\indivWom), which the instruction does not address, persist. A system-prompt
fix thus changes which groups appear homogeneous without removing the register
or the contrast that is strongest in pretrained models.

\section{Template versus Prompt}
\label{app:template}

The chat condition differs from the post-trained condition in two ways at
once: the post-trained model sits behind its chat template with the system
message \textit{``You are a chatbot''} (the system message of the original
study), and it receives the original
instruction instead of the few-shot frame. Two further conditions change one
at a time, on the same post-trained checkpoints and at the same decoding
setting. \textit{Few-shot in template} gives the frame's three exemplars as
prior user and assistant turns (\textit{``a thirty-word story about a
firefighter.''} followed by its text), then the target request, all behind
the chat template with the system message. A first version that placed the
whole frame in one user turn was abandoned: Llama-3.1 rewrote one of the
exemplars in 42\% of its replies. \textit{Instruction, no template} gives the
original instruction as plain text.

\paragraph{Cleaning.} Without its template, a post-trained model often opens
with a heading (\textit{``Here is a thirty-word story about \dots:''}), which
we keep with the paragraph it introduces, as the chat condition keeps it.
Some replies instead continue the instruction (\textit{``Answer in
English.''}) or write out reasoning (\textit{``Okay, the user wants
\dots''}); these are not texts about a person and are removed like refusals.
They are under 5\% of replies in six families and 16\% in Qwen3, where they are
more frequent for White subjects (24\%) than for others (12--15\%). Excluding
Qwen3 leaves every conclusion below unchanged.

\begin{table*}[t]
\centering
\small
\begin{tabular}{@{}llrrrrr@{}}
\toprule
Template & Prompt & African Am. & Asian Am. & Hispanic Am. & Woman & ID min./White \\
\midrule
no & few-shot & $+0.01$ (0.00) & $+0.12$ (0.08$^{\dagger}$) & $+0.39$ (0.08$^{\dagger}$) & $+0.19$ (0.04) & 1.4 / 0.4 \\
yes & few-shot & $+0.02$ (0.12$^{\dagger}$) & $+0.09$ (0.05) & $+0.48$ (0.15$^{\dagger}$) & $+0.16$ (0.05) & 2.8 / 0.7 \\
no & instruction & $+0.02$ (0.08$^{\dagger}$) & $+0.06$ (0.08$^{\dagger}$) & $+0.33$ (0.07) & $+0.19$ (0.09$^{\dagger}$) & 2.6 / 0.7 \\
yes & instruction & $-0.08$ (0.13$^{\dagger}$) & $-0.04$ (0.15$^{\dagger}$) & $+0.38$ (0.11$^{\dagger}$) & $+0.17$ (0.03) & 4.5 / 1.2 \\
\bottomrule
\end{tabular}
\caption{Template $\times$ prompt on the post-trained model, seven families.
Pooled random-effects contrast (SD units, group term masked) with
between-family $\tau$ in parentheses; $^{\dagger}$Cochran's $Q$ $p<.05$.
\textit{ID}: identity-lexicon words per 100 for minority / White subjects.
The first row is \textit{Post-trained}, the last \textit{Chat}.}
\label{tab:template}
\end{table*}

\paragraph{Results.} Each ingredient contributes (Table~\ref{tab:template}).
The template alone raises the African American $\tau$ to \tmplTauAfr, close to
the chat condition's \chatTauAfr{} ($Q$ on the change from the pretrained model,
$p = \tmplDeltaQpAfr$), but leaves the Asian American $\tau$ at \tmplTauAsn{}
($p = \tmplDeltaQpAsn$);
the instruction alone raises both to \rawiTauAfr{} and \rawiTauAsn. The pooled
contrasts stay at or above zero in both, and turn negative only when the two
are combined. The identity vocabulary behaves the same way: each ingredient
alone roughly doubles it for minority subjects, and the combination raises it
further. The African American spread thus arises mostly from the dialogue format. The
sign change in both contrasts, and the Asian American spread, appear only
when the template and the original request are combined. Both conditions also
make the models name the person more often than the post-trained version does
(\namedTmplMin--\namedTmplMax\% of texts in the template condition and
\namedRawiMin--\namedRawiMax\% in the instruction condition, against
\namedInstMin--\namedInstMax\%), which fits the role of names in
Section~\ref{sec:names}.

\section{Hiding Names and Places}
\label{app:names}

After the group term is replaced, an automatic tagger (spaCy~3.8 with
\texttt{en\_core\_web\_lg}) finds named entities in every text. Each person's
name is replaced by \texttt{[NAME]}, and each place (countries, cities, states
and other locations) by \texttt{[PLACE]}. The same placeholders are used for
every group, so hiding a name removes the name's content but keeps the fact
that a name was given. The texts, filters and estimator are otherwise those
of the main analysis, at temperature 1 and top-$p$ 1.

\paragraph{Which entities matter.} Hiding places alone changes no
conclusion: the families' changes in chat still differ for African and Asian
Americans ($p=\varPlChatDeltaQpAfr$ and \varPlChatDeltaQpAsn), and the
seven chat models' pooled Hispanic effect is \varPlPoolHisp~SD. Hiding names
and places together gives the same results as hiding names alone
($p=\varNmPlChatDeltaQpAfr$ and \varNmPlChatDeltaQpAsn; Hispanic
\varNmPlPoolHisp~SD). Hiding nationality and language words as well
(\textit{Mexican}, \textit{Spanish}) leaves the Hispanic effect at
\varCulPoolHisp~SD. The Hispanic effect is therefore not carried by names,
places or words of national origin.

\paragraph{Names the tagger misses.} Checked against 26 common first names, the
tagger misses 1--3\% of their occurrences in chat texts about White, African
American and Asian American subjects and 6\% in texts about Hispanic subjects.
The most frequent tags that are not names (for example \textit{Avid} in
``avid reader'', \textit{baker} and \textit{abuela}) make up about 2\% of all
tags. A second pass hides
every word that the tagger marked as a person's name at least ten times
elsewhere. It leaves the results unchanged: the families' changes in chat do
not differ for African Americans ($p=\dictChatDeltaQpAfr$), and the Hispanic
effect of the seven chat models is \dictPoolHisp~SD.

\paragraph{Second encoder.} With \texttt{all-MiniLM-L12-v2}, the families'
changes in chat differ for African Americans with only the group term hidden
($p=\miniChatDeltaQpAfr$) but not with names hidden ($p=\miniNmChatDeltaQpAfr$).
The seven chat models' Hispanic effect falls from \miniPoolHisp{} to
\miniNmPoolHisp~SD with names hidden, positive in \miniNmPosHisp{} of seven
models, and its 95\% interval for a new model, \miniNmPIHisp, stays above
zero. The gender effect is weaker under this encoder whether or not names are
hidden: \miniPoolWom~SD (positive in \miniPosWom{} of seven models, interval
\miniPIWom) with the group term hidden, and \miniNmPoolWom~SD (positive in
\miniNmPosWom{} of seven) with names hidden as well.

\paragraph{Name reuse by model.} Table~\ref{tab:reuse} gives, for each chat
model and group, the share of texts naming the person whose first name is one
of that model's five most common names for the group.

\begin{table}[h]
\centering
\small
\begin{tabular}{@{}lrrrr@{}}
\toprule
Model & White & African & Asian & Hispanic \\
\midrule
Llama-3.1 & 70 & 28 & 40 & 63 \\
Qwen3 & 60 & 81 & 66 & 87 \\
Falcon3 & 46 & 42 & 48 & 49 \\
Mistral-NeMo & 48 & 33 & 53 & 63 \\
OLMo-2 & 27 & 36 & 35 & 70 \\
Granite-3.3 & 57 & 31 & 28 & 79 \\
gpt-oss & 41 & 66 & 43 & 67 \\
\bottomrule
\end{tabular}
\caption{Name reuse in the chat models: percentage of texts that name the
person whose first name is one of the model's five most common names for that
group.}
\label{tab:reuse}
\end{table}

\section{Statistical Notes}
\label{app:stats}

\paragraph{Contrasts.} The cell-level model includes the race $\times$ gender
interaction. Each race contrast is the difference from White subjects averaged
over the two genders, and the gender contrast is the difference from men
averaged over the four race groups; both are linear combinations of the
interaction model's coefficients.

\paragraph{Placebo.} For each chat model and sampled decoding setting, each of
the 26 White-subject cells is split at random into two halves of 25 texts. The
pairwise estimator fits
$\text{cosine} \sim \text{half} \times \text{gender} + (1 \mid \text{format})$ to
all within-half pairs; the cell-level estimator fits
$\text{MPCS} \sim \text{half} \times \text{gender} + \text{format}$ to the 52
half-cells with HC3 errors. Three splits per setting give 420 tests per
estimator. The pairwise estimator rejects at $\alpha=.05$ in \placeboRate\% of
them, the cell-level estimator in \placeboCell\%.

\paragraph{Original corpus.} On the GPT-3.5 corpus of \citet{lee_large_2024b},
the cell-level estimator recovers all three positive race contrasts ($+0.28$,
$+0.26$, $+0.22$ SD with group terms masked), so the original finding is not an
artifact of the pairwise estimator. The original study's smaller gender contrast is not
significant under the cell-level estimator ($+0.11$ SD with group terms masked,
$p=.08$). Subsampled to 50 texts per cell over 50 draws, the three raw race contrasts are
significant in every draw with a standard deviation of 0.03 across draws. Fifty
texts per cell is enough: the uncertainty in a cell-level estimate comes from
variation between prompt cells, not from the number of texts within them.

\paragraph{Heterogeneity.} $\tau$ is the DerSimonian--Laird estimate. With seven
families, Cochran's $Q$ has limited power. Simulating seven estimates with the
pretrained models' standard errors, $Q$ at $\alpha=.05$ detects a true spread
of $\tau=0.10$ SD with probability \powTenAfr\% (African American) and
\powTenAsn\% (Asian American), and a spread of 0.15, the chat level, with
\powFifteenAfr\%. A non-significant $Q$ for the pretrained models therefore
makes a spread as large as chat's unlikely, but does not exclude a smaller
one. It is the
contrast with the chat condition, in which the same test on the same families
rejects, that carries the argument. Q-profile 95\% intervals
\citep{viechtbauer_confidence_2007} for $\tau$ are \baseTauCIAfr{} and
\baseTauCIAsn{} for the pretrained African and Asian American contrasts and
\chatTauCIAfr{} and \chatTauCIAsn{} in chat: the chat spread excludes zero,
the pretrained spread does not, and the intervals overlap. A direct test that
the chat spread exceeds the pretrained spread enumerates all $2^7$ within-family
swaps of the two conditions' estimates, each condition first centered on its
pooled value; it gives $p=\swapPAfr$ and \swapPAsn, with a smallest attainable
$p$ of $1/128$. A parametric bootstrap that redraws each family's estimate from its sampling
distribution, conditional on these seven families, gives 95\% intervals for
$\tau_{\text{chat}}-\tau_{\text{pre}}$ of \bootCIAfr{} and \bootCIAsn{}
(one-sided $p=\bootPAfr$ and \bootPAsn). Against the post-trained condition,
the swap test gives $p=\swapInstPAfr$ and \swapInstPAsn. Our cell-level tests are conservative (\placeboCell\% placebo rejections),
which makes $Q$ less likely to reject in every condition, chat included; the
power simulation above uses the same standard errors. We therefore rest the
argument on the change test, which asks
the more specific question whether families move by different amounts. Chat
estimates are less precise than pretrained ones (median standard error
\medSeChat{} against \medSeBase), which lowers rather than raises the
estimated chat spread. The test on the change from
pretrained to chat (\textit{$Q$ on $\Delta$}) uses each family's difference with
its standard error $\sqrt{\text{se}_{a}^2 + \text{se}_{b}^2}$, which treats the
two conditions' estimates as independent; they are, since the texts are sampled
separately.

\end{document}